\documentclass[runningheads]{llncs}
\usepackage[T1]{fontenc}
\usepackage{graphicx,verbatim}
\usepackage{multirow}
\usepackage{booktabs}
\usepackage{amssymb}
\usepackage{amsmath}
\begin{document}
\title{Beyond Rigid AI: Towards Natural Human-Machine Symbiosis for Interoperative Surgical Assistance}

\author{
Lalithkumar Seenivasan\inst{1,\star} \and       
Jiru Xu\inst{1,\star} \and            
Roger D. Soberanis Mukul\inst{1} \and
Hao Ding\inst{1} \and
Grayson Byrd\inst{1,2} \and
Yu-Chun Ku\inst{1} \and
Jose L. Porras\inst{3} \and
Masaru Ishii\inst{3} \and
Mathias Unberath\inst{1}
}
\authorrunning{Seenivasan et al.}
%
\institute{Johns Hopkins University, Baltimore MD, USA \and
Johns Hopkins Applied Physics Laboratory, Laurel MD, USA \and
Johns Hopkins Medical Institutions, Baltimore MD, USA   \\
\email{\{lseeniv1,unberath\}@jhu.edu}\\
$\star$ Equal contribution
}

\maketitle              
\begin{abstract}
Emerging surgical data science and robotics solutions, especially those designed to provide assistance in situ, require natural human-machine interfaces to fully unlock their potential in providing adaptive and intuitive aid. Contemporary AI-driven solutions remain inherently rigid, offering limited flexibility and restricting natural human-machine interaction in dynamic surgical environments. These solutions rely heavily on extensive task-specific pre-training, fixed object categories, and explicit manual-prompting. This work introduces a novel Perception Agent that leverages speech-integrated prompt-engineered large language models (LLMs), segment anything model (SAM), and any-point tracking foundation models to enable a more natural human-machine interaction in real-time intraoperative surgical assistance. Incorporating a memory repository and two novel mechanisms for segmenting unseen elements, Perception Agent offers the flexibility to segment both known and unseen elements in the surgical scene through intuitive interaction. Incorporating the ability to memorize novel elements for use in future surgeries, this work takes a marked step towards human-machine symbiosis in surgical procedures. Through quantitative analysis on a public dataset, we show that the performance of our agent is on par with considerably more labor-intensive manual-prompting strategies. Qualitatively, we show the flexibility of our agent in segmenting novel elements (instruments, phantom grafts, and gauze) in a custom-curated dataset. By offering natural human-machine interaction and overcoming rigidity, our Perception Agent potentially brings AI-based real-time assistance in dynamic surgical environments closer to reality.



\end{abstract}

\section{Introduction} \label{Section:introduction}
Surgical procedures demand seamless coordination between surgical teams and technology to ensure efficiency, and improved patient outcomes~\cite{guni2024artificial}. With surgical procedures becoming increasingly complex, the need for intelligent assistance systems that can naturally integrate and adapt to emerging dynamic surgical workflows and intraoperative conditions has become critical. While recent advancements in surgical data science and robotics have accelerated the development of AI-driven solutions for intraoperative assistance~\cite{ding2024towards,shu2023twin}, these solutions remain fundamentally rigid. Most contemporary AI-based solutions operate within predefined tasks and classes, lacking the flexibility to engage in natural human-machine collaboration. To truly transform surgical workflows, intelligent systems must move beyond rigid, pre-defined task execution to enable real-time and natural human-machine interaction. A flexible system offering natural human-machine interaction and easy integration into existing workflow with minimal disruption will enhance decision-making and procedural efficiency, leading to better patient outcomes.

With the long-term vision of enabling intelligent intraoperative surgical assistance, numerous AI-driven solutions have been proposed for surgical scene analysis, including semantic segmentation~\cite{seenivasan2022global,ding2024segstrong,ding2022carts,ding2023rethinking,yue2024surgicalsam,shen2024performance}, surgical scene graphs~\cite{islam2020learning,seenivasan2022global,holm2023dynamic}, triplet detections~\cite{nwoye2022rendezvous,gui2023mt4mtl,xi2023chain}, and surgical phase detections~\cite{ding2024towardsSPR,ding2024neural,yang2024surgformer}. However, the rigidity of these current solutions limits their integration into routine surgical workflows. With regard to the segmentation task, current models are limited to segmenting only predefined surgical elements (instruments and tissues) and lack the flexibility to segment novel elements without additional training. Although recent foundation models, such as the Segment Anything Model (SAM)~\cite{ravi2024sam}, show potential to address these concerns, rigidity still remains in their requirement for manual prompting. 
This need for manual prompting prevents seamless, intuitive interaction with surgeons and disrupts the normal surgical workflow. 
Achieving real-time segmentation of novel elements through natural human-machine interaction is critical to realizing the full potential of AI-based surgical assistance. In vascular and reconstructive surgeries, real-time segmentation and tracking of custom-shaped grafts could assist surgeons in ensuring proper positioning, monitoring shape integrity, and detecting unintended shifts, ultimately improving anastomotic precision, reducing misalignment risks, and enhancing overall surgical outcomes \cite{venugopal2022real,wu2011segmentation}. 
Similarly, intelligent tracking of surgical sponges and gauze could help prevent retained foreign objects, reducing postoperative complications~\cite{de2020automatic,sanchez2022gauze}. 

To address the limitations associated with the rigidity of contemporary AI models and enable natural human-machine partnership in surgery, we introduce Perception Agent —a novel AI-driven system capable of real-time, on-demand segmentation of surgical elements, leveraging Large Language Models (LLMs), SAM2, and any-point tracking foundation models. Incorporating a memory repository and two novel approaches for segmenting novel elements - (i) object-centric and (ii) reference-based segmentation- our agent, unlike existing works, offers the flexibility to segment both known and novel elements in the surgical scene through intuitive hands-free human-machine interaction that naturally integrates with the surgical workflow. Additionally, it incorporates the capability to store (learn) the memory of novel elements in real-time for use in future surgeries, marking this work as an early step toward human-machine symbiosis, where the machine not only assists humans but also continuously learns and adapts to emerging surgical elements through human-machine interaction.

\section{Methodology}

\subsection{Preliminaries}

\noindent\textbf{(a) Segment Anything Model 2 (SAM2)~\cite{ravi2024sam}:}
SAM2 is a prompt-able segmentation foundation model. Like SAM~\cite{kirillov2023segment}, its predecessor, it employs an image encoder, a prompt encoder, and a mask decoder to generate segmentation masks based on the input prompt. Additionally, SAM2 introduces memory blocks to retain knowledge from previous outputs, expanding its segmentation capabilities to video. The memory attention module and memory encoder are used to generate memory embeddings which are stored in a First-in-first-out (FIFO) queue within the memory bank, which retains $n$ recent frames as history. Additionally, the memory bank maintains user-prompted frame information in a separate FIFO queue for $m$ user-prompted frames.

\noindent\textbf{(b) CoTracker3~\cite{karaev2024cotracker3}:}
CoTracker3 is a transformer-based foundation model for dense point tracking across video sequences. Given a set of query points ($Q\in \mathbb{R}^3$) in a query frame, it estimates their corresponding positions across subsequent video frames. In addition to tracking points, it estimates point occlusion and uncertainty for the generated tracks. In the online mode, the model operates in a sliding window approach (window size = $16$), offering continuous tracking.


\subsection{Perception Agent}

\begin{figure}[!t]
  \centering
  \includegraphics[width=0.95\linewidth]{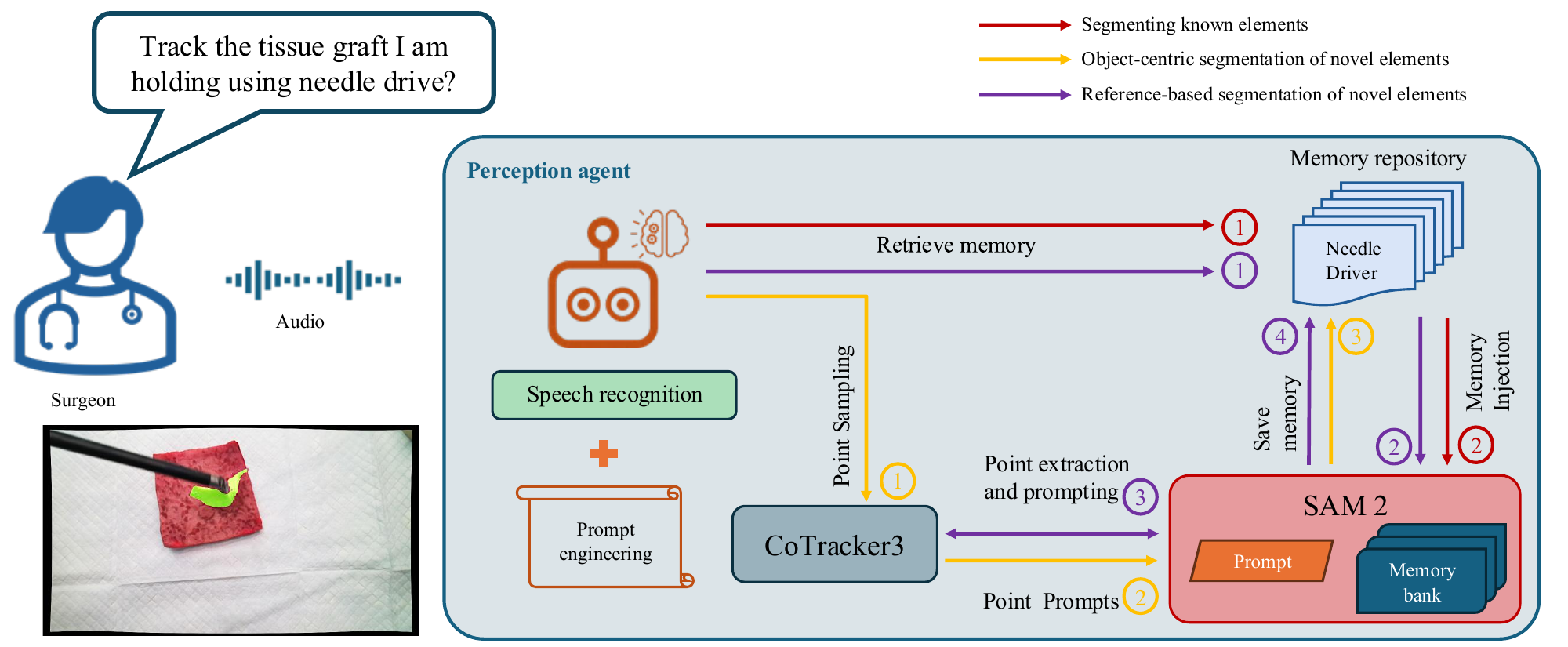}
  \caption{Perception Agent enabling natural human-machine interaction for on-demand segmentation of known and novel surgical elements, by levering foundation models and incorporating memory repositories.}
  \label{fig:method}
\end{figure}

We introduce Perception Agent, an AI-driven system that facilitates natural human-machine interaction for on-demand segmentation of surgical elements (Fig~\ref{fig:method}). Integrating speech-incorporated LLMs (Whisper and GPT-4o), any-point tracking (CoTracker3), and segmentation (SAM2) foundation models, and incorporating a memory repository $M = \{m_1, m_2, ..., m_n\}$ that holds the memory of known surgical instruments, the agent offers a holistic and flexible solution for segmenting both known and previously unseen elements. While the agent primarily employs SAM2 to generate segmentation masks, it leverages an LLM's flexibility to (i) interpret the surgeon's natural language instructions, (ii) segment known elements by coordinating between the memory repository and SAM2, and (iii) segment novel elements through two distinct mechanisms: object-centric approach, which integrates CoTracker3 and SAM2, and reference-based tracking, which coordinates the memory repository, CoTracker3, and SAM2. By enabling speech-based interaction and utilizing any-point tracking for hands-free, motion-based prompting to segment novel elements, the agent facilitates natural interaction with surgeons in the surgical setting.

\noindent\textbf{(a) Task and Element Extraction:} With speech being the primary mode of interaction, the surgeon's casual audio instructions are transcribed into text and processed by the Perception Agent. Here, the agent is prompt-engineered to extract the task $T = \{start~tracking,~stop~tracking\}$ and identify the element to be segmented. The subsequent action of the agent is prompt-engineered to depend on whether a prior memory of the element exists and if the object-centric (e.g., track the needle drive) or reference-based (e.g., track the tissue I am holding using the needle driver) segmentation approach is required. 

\noindent\textbf{(b) Segmenting Known Element:}
When the agent is tasked with tracking an element, the agent first queries for corresponding memory from its memory repository. Here, the agent tries to match the casual name provided by the surgeon (e.g., needle driver) to the exact stored name in the memory repository (e.g., large needle driver). If a matching memory is found, the agent retrieves the memory embedding of the element and injects it into the active SAM2 session, allowing immediate object tracking. 

\begin{figure}[!t]
  \centering
  \includegraphics[width=.95\linewidth]{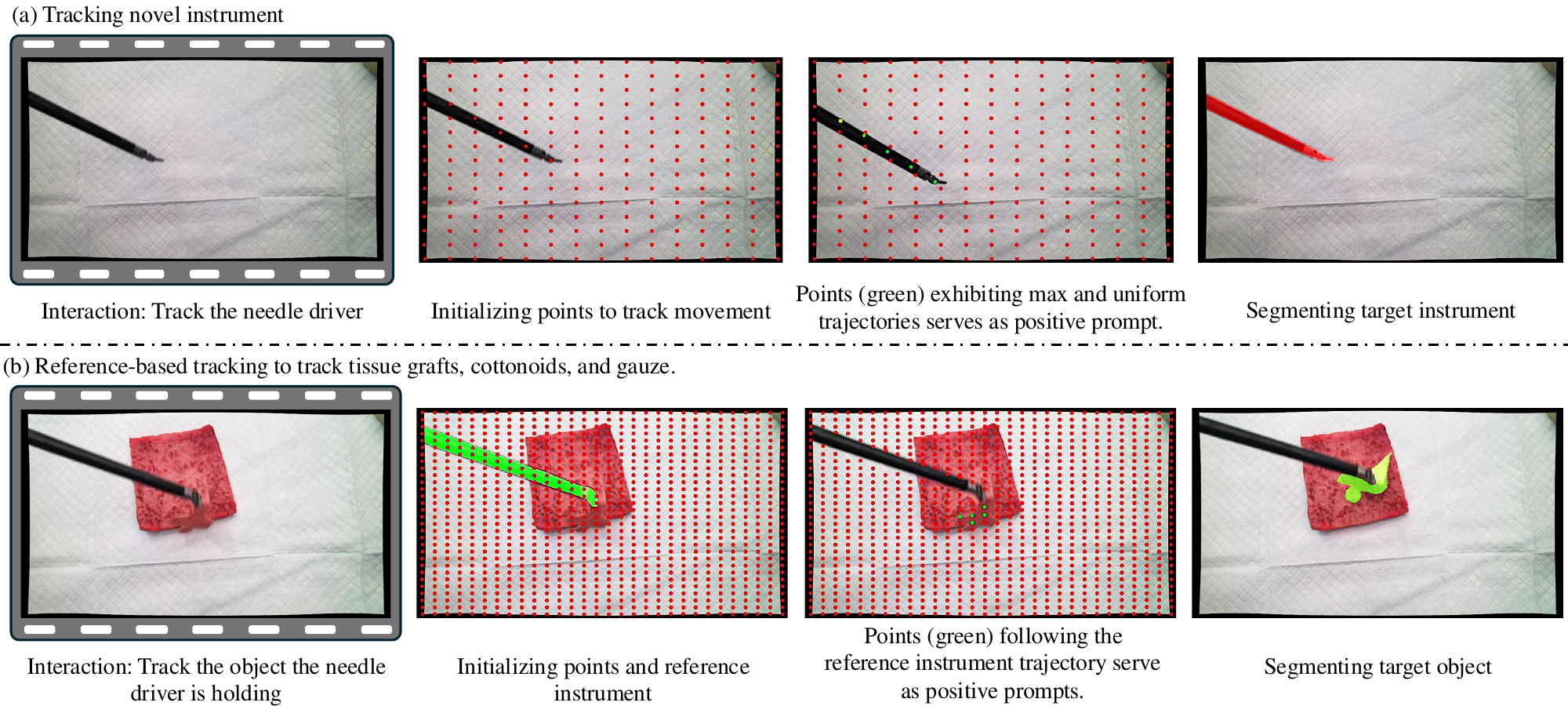}
  \caption{(i) Object-centric and (ii) reference-based segmentation of novel surgical elements in the surgical scene.}
  \label{fig:novel_tracking}
\end{figure}

\noindent\textbf{(c) Object-Centric Segmentation of Novel Element:}
When tasked with tracking an instrument for which the agent doesn't have a corresponding memory in its memory repository, the agent initiates an object-centric tracking routine (Fig~\ref{fig:novel_tracking}(a)). Firstly, the surgeon is notified of the initiation of the tracking process. Secondly, dense query points ($Q_p$) are populated across the surgical scene and tracked using CoTracker3 across $16$ frames. To identify the novel instrument, we filter points exhibiting the most significant and uniform motion across consecutive frames -- corresponding to the surgeon manipulating the novel instrument. Let $Q_i(t)$ denote the $i^{th}$ point at $t$ ($t= 0,1,2,...,T$). We calculate the motion $V_i(t)$ for all points, the total displacement of each point $D_i$, and the index $i^*$ of the point with the largest displacement (reference point) based on:

\begin{equation}
    V_i(t) = Q_i(t) - Q_i(t - 1);~D_i = \sum_{t=1}^T \| V_i(t) \|_2; ~i^* = \arg \max_{i=0,1,...} D_i \label{eq:motion}
\end{equation}
We then calculate the average cosine similarity for all points with $Q_{i^*}$:
\begin{equation}
    S_i = \frac{\sum_{t=1}^T \frac{V_i(t) \cdot V_{i^*}(t)}{\| V_i(t) \| \|V_{i^*}(t) \|}}{T} \label{eq:sim}
\end{equation}
and apply a threshold $\gamma$ to filter the non-matching points $\mathcal{I} = \{i \mid S_i > \gamma \}$. Lastly, we find the Top $k$ matching points $\mathcal{I}^{\prime} \subset \mathcal{I}$ if $k < | \mathcal{I} |$. The resulting point set $Q^* = \{ Q_i \mid i \in \mathcal{I}^{\prime} \}$ is injected as point prompts into the SAM2, enabling segmentation of the novel instrument. Simultaneously, the novel instrument's memory is stored inside the memory repository for future retrieval and tracking.

\noindent\textbf{(d) Reference-based Segmentation of Novel Element:}
The agent also supports a reference-based segmentation of novel elements, allowing tasks such as "tracking the tissue that the needle driver is holding". In this approach, first, a set of query points ($Q_p$) are populated across the initial scene and tracked by CoTracker3 across $48$ frames. Simultaneously, the reference object (e.g., needle driver) is segmented by SAM2 using the memory from the memory repository, across the same $n$ frames. The trajectories of the $Q_p$ are then analyzed and filtered based on two criteria: (i) A displacement threshold to prune static points and (ii) A cosine similarity measure that compares the trajectories of $Q_p$ against the trajectories of points within the segmentation mask of the reference object, to prune points with mismatched motion pattern. Let $Q_i(t)$ denote the $i^{th}$ point at $t$ ($t=0,1,2,...,T$). We first separate the points into $R(t) \subset Q(t)$ and $C(t) \subset Q(t)$, where for all $i$ and $t$, $R_i(t) \in R(t)$ and $C_i(t) \in C(t)$ lies inside and outside of the reference object's segmentation mask, respectively. Similar to \eqref{eq:motion}, we find motion $V^R$ and $V^C$ for the reference points and candidate points. Then, we calculate the motion template of the reference points ($V^*(t)$), and calculate the average cosine similarity of all points ($S_i$) in the candidate set with the motion template based on:
\begin{equation}
    V^*(t) = \frac{\sum_i R_i(t)}{|R(t)|}; S_i = \frac{\sum_{t=1}^T \frac{V_i^C(t) \cdot V^*(t)}{\| V_i^C(t) \| \|V^*(t)\|}}{T}
\end{equation}
we then apply a threshold $\gamma$, similar to the object-centric approach, to filter point set $Q^*$ that is then used to prompt SAM2 to segment the target element.

\section{Experiment and Results}
SAM has been widely validated on surgical datasets~\cite{shen2024performance,yue2024surgicalsam,soberanis2024gsam+}. 
As our primary contribution is a flexible perception agent that allows on-demand segmentation of surgical elements through natural human-machine interaction, and not an end-to-end model, instead of benchmarking against state-of-the-art models, we carefully design the experiment to benchmark our agent's flexibility in (i) segmenting known elements from public EndoVis18~\cite{allan20202018}) dataset and (ii) novel elements on in-house dataset generated using the da Vinci Research Kit 
(dVRK)~\cite{kazanzides2014open}.

\begin{table}[!b]
\centering
\caption{Quantitative studies on segmentation performance based on prior memory using manual-prompting strategy vs our perception agent's object-centric approach.}
\scalebox{0.75}{
\begin{tabular}{l|c|c|c|c|c|c|c|c|c}
\toprule
\multicolumn{1}{c|}{\multirow{2}{*}{\textbf{Methods}}} & \multicolumn{8}{c|}{\textbf{Dice score}}                                        & \multirow{2}{*}{\textbf{mIoU}} \\ \cline{2-9}
\multicolumn{1}{c|}{}                         & \textbf{T1}     & \textbf{T2 }    & \textbf{T3}     & \textbf{T4}     & \textbf{T5}     & \textbf{T6}     & \textbf{T7}     & \textbf{Avg}    &                       \\
\midrule
Manual-prompting                       & 0.987 & 0.956 & 0.919 & 0.991 & 0.949 & 0.990 & 0.951 & 0.963 & 0.621                \\
\midrule
\multicolumn{10}{c}{Comparision study}                                                                              \\
\midrule
Prior-memory from Manual-prompting  & 0.987 & -      & 0.919 & 0.991 & -      & -      & -      & 0.965 & 0.738                \\
Prior memory from Object-centric approach  & 0.976 & -      & 0.924 & 0.991 & -      & -      & -      & 0.963 & 0.715  \\
\bottomrule
\end{tabular}}
\label{Table:1}
\end{table}

\begin{figure}[!b]
  \centering
  \includegraphics[width=0.85\linewidth]{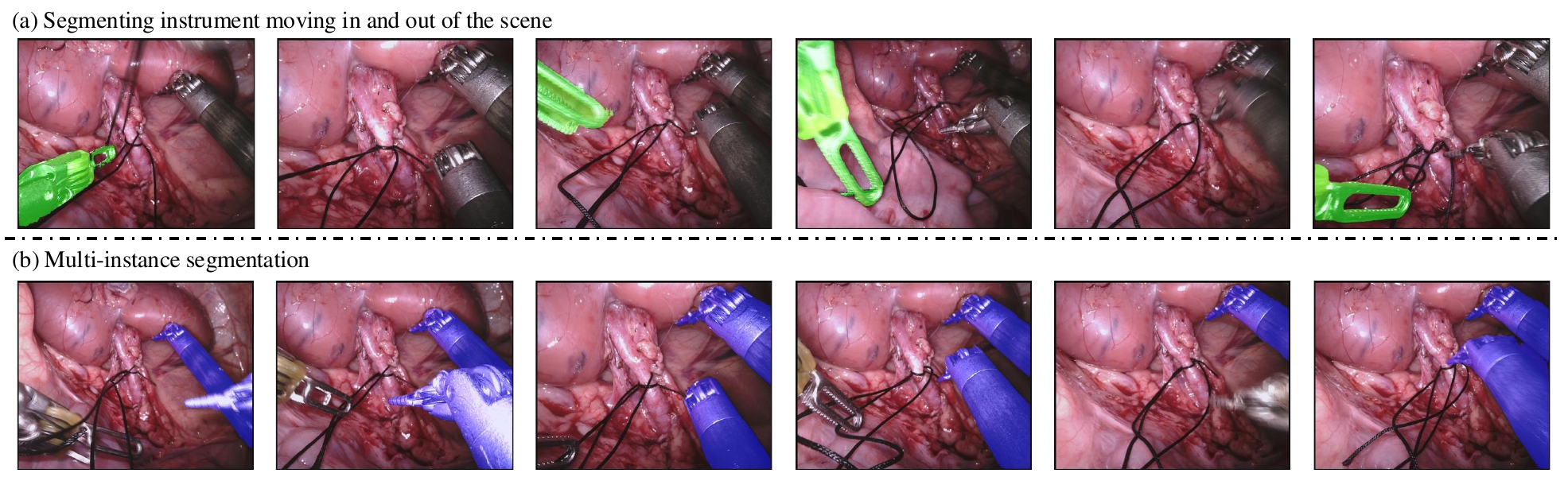}
  \caption{Qualitative analysis of our Perception Agent in segmentation (i)instruments moving in and out of the surgical scene and (ii) multiple instances of the same instrument based on prior memory.}
  \label{fig:in_out_multi_instance}
\end{figure}

\begin{table}[!t]
\centering
\caption{Quantitative analysis of agent's performance when using memory from previous surgery vs same surgery.}
\scalebox{0.85}{
\begin{tabular}{c|c|c|c|c|c|c|c}
\toprule
\multirow{2}{*}{\textbf{Dataset}} & \multirow{2}{*}{\textbf{Memory}} & \multicolumn{5}{c|}{\textbf{Dice score}}     & \multirow{2}{*}{\textbf{mIoU}} \\
                                  &                                  & \textbf{T1}     & \textbf{T3}     & \textbf{T4}     & \textbf{T5}    & \textbf{Avg}     &                                \\
\midrule
EndoVis18                         & EndoVis18                        & 0.987 & 0.919 & 0.991 & 0.949 & 0.961 & 0.698                       \\
EndoVis18                         & Endovis17                        & 0.961 & 0.947  & 0.979 & 0.916 & 0.951  & 0.625 \\
\bottomrule
\end{tabular}}
\label{Table:between_surgery_memory}
\end{table}

\subsection{Quantitative and Qualitative Analysis on Public Dataset:}
We quantitatively analyze our agent's performance in segmenting surgical elements based on prior memory generated using (i) manual-prompting and (ii) agent's object-centric approach (Table~\ref{Table:1}). For this study, we use $16$ video sequences from the EndoVis18 dataset~\cite{allan20202018}. Three sequences (1, 5, $\&$ 16) were used as the held-out test set. The remaining sequences served as the training set, from which, $1$ frame was utilized for creating a memory for each instrument. Firstly, from Table~\ref{Table:1}, we observe that our Perception Agent achieved an average (avg) Dice score of 0.963 and a mean Intersection over Union (mIoU) of 0.621 when utilizing prior memory generated using manual-prompting on cherry-picked frame (the instrument was clearly visible and well-define) from the training set. Secondly, our agent's segmentation performance based on prior memory generated using our object-centric approach is benchmarked against using prior memory generated using manual-prompts. Since only three instruments (bipolar forceps, large needle driver / monopolar curved scissors) met the criteria -- the only instrument having significant motion in the $16$ consecutive frames -- in the available train set to employ our object-centric approach, we benchmarked both approaches on segmenting these three instruments in the test set. We observe that segmentation based on memory created using an object-centric approach performed similarly (average Dice score) to when using prior memory created from manual-prompting. 
Qualitatively (Fig.~\ref{fig:in_out_multi_instance}), we observe that our agent can fully exploit the SAM2 model's capabilities in (i) tracking instruments moving in and out of the scene, and (ii) tracking multiple instances of the same instruments, based on prior memory of the instruments stored in our memory repository. The perception agent's flexibility in segmenting instruments based on memory from previous surgery -- mimicking human's learning ability -- is also quantitatively analyzed in Table~\ref{Table:between_surgery_memory}. Here, simulating two surgeries, the memory of an instrument generated from a frame in EndoVis17~\cite{allan20192017} dataset is used as prior knowledge in segmenting the instrument in the EndoVis18~\cite{allan20202018} test set. We observe that, the performance of the agent in segmenting standard tools based on memory from previous surgery is similar (Average Dice score) to performance when memory is extracted from very similar sequences.


\subsection{Qualitative Analysis on Simulated Surgical Application}

\begin{figure}[!t]
  \centering
  \includegraphics[width=0.9\linewidth]{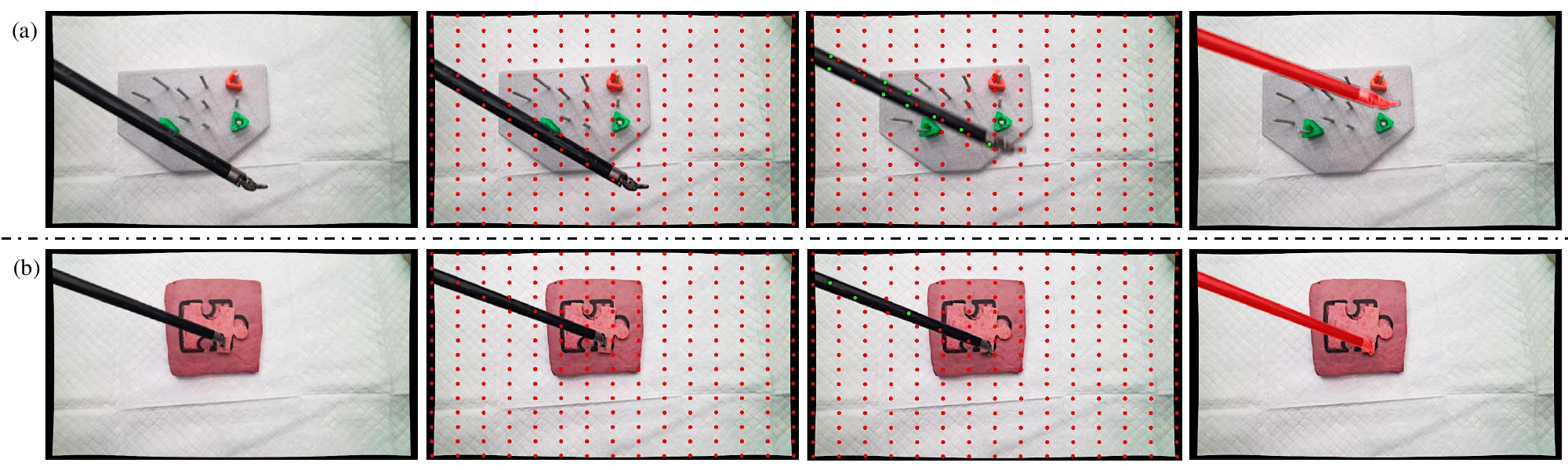}
  \caption{Qualitative analysis of Perception agent's object-centric approach in segmenting novel instruments with (i) peg-board and (ii) phantom tissue background.}
  \label{fig:dvrk_instrument_tracking}
\end{figure}

\begin{figure}[!t]
  \centering
  \includegraphics[width=1.0\linewidth]{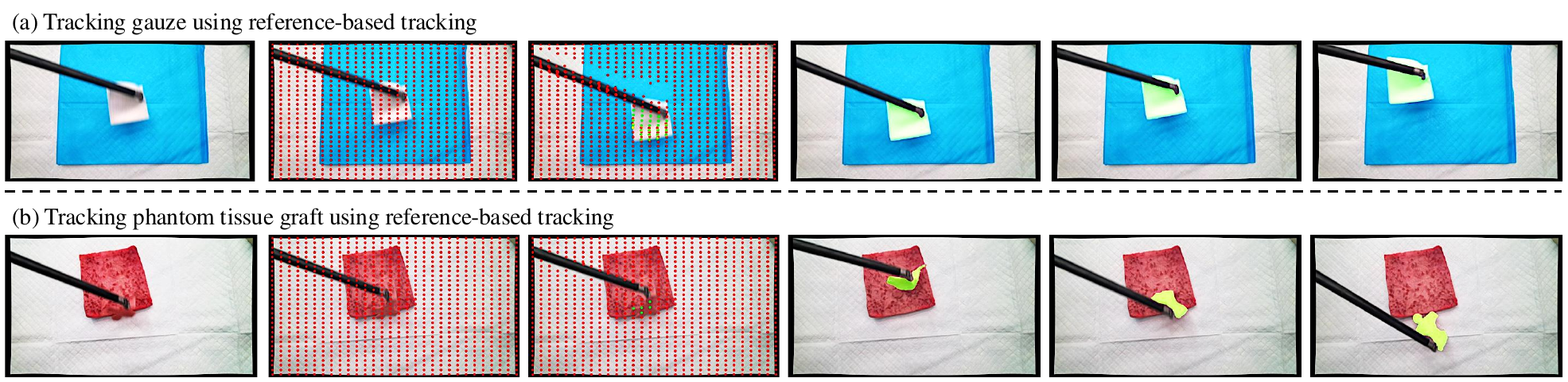}
  \caption{Qualitative analysis of Perception agent's reference-based segmentation approach in segmenting (gauze) and phantom tissue grafts.}
  \label{fig:dvrk_target_tracking}
\end{figure}

Our agent's flexibility in segmenting novel surgical elements using (i) object-centric and (ii) reference-based approaches are analyzed qualitatively in a simulated surgical setting, using videos generated from dVRK. To validate the object-centric approach, two environments -- one with a peg-board background and another with a phantom tissue background -- were used to curate the video. From Fig.~\ref{fig:dvrk_instrument_tracking}, we observe that the agent can successfully segment the novel instrument based on the object-centric approach. For validating reference-based tracking, we also simulate two environments - one to segment the gauze held by the needle driver, and the other to segment a phantom tissue with a phantom tissue background. Fig.~\ref{fig:dvrk_target_tracking} shows our agent's flexibility in successfully segmenting the gauze and phantom tissue graft using the reference-based approach.

\section{Discussion and Conclusion}

In this work, we introduce Perception Agent, an AI-driven early prototype that integrates speech-enabled LLMs, Segmentation, and any-point tracking foundation models to provide real-time intraoperative surgical assistance through natural human-machine interaction. Leveraging these foundation models and further incorporating a memory repository along with two novel motion-based prompting mechanisms, the agent overcomes the rigidity of contemporary AI solutions, enabling on-demand segmentation of both known and unseen surgical elements. By integrating the ability to segment novel elements and retain its memory for future surgery through intuitive interaction with surgeons, our perception agent marks a foundational step towards human-machine symbiosis. 

While our Perception Agent marks an early step toward realizing real-time, adaptive AI assistance in surgery, challenges remain, particularly in tracking very small objects with minimal feature information, and robust performance in segmenting multiple instances of similar instruments. Additionally, segmenting unseen static surgical elements remains a challenge due to the lack of movement cues. Refining the system’s memory update mechanism with a structured, hierarchical approach would further enhance long-term retention of clinically relevant elements while optimizing computational efficiency. Beyond its immediate clinical applications, Perception Agent also has the potential to play a key role in enabling multi-agent-driven surgical automation, where perception, planning and action agents could collaboratively work in automating surgical tasks. It may also be employed for rapid pseudo-annotation of datasets for developing downstream applications.

\bibliographystyle{splncs04}
\bibliography{references}

\end{document}